\documentclass[10pt,journal,compsoc]{IEEEtran}

\usepackage{epsfig} 
\usepackage{mathptmx} 
\usepackage{amsmath} 
\usepackage{multirow}
\usepackage{float}
\usepackage{graphicx}
\usepackage{hyperref}
\ifCLASSOPTIONcompsoc
  \usepackage[nocompress]{cite}
\else
  \usepackage{cite}
\fi

\begin{document}

\title{Impact of Face Image Quality Estimation on Presentation Attack Detection}
%
%

\author{Carlos~Aravena,
        Diego~Pasmino, Juan~E.~Tapia ~\IEEEmembership{Member,~IEEE,}
        and~Christoph~Busch,~\IEEEmembership{Senior Member,~IEEE}
\IEEEcompsocitemizethanks{\IEEEcompsocthanksitem Carlos Aravena and Diego Pasmino are with R\&D Center, TOC Biometrics, Santiago, Chile.\\

\IEEEcompsocthanksitem Juan Tapia and Christoph Busch, da/sec-Biometrics and Internet Security Research Group, Hochschule Darmstadt, Germany.
\protect\\ %
Corresponding author: juan.tapia-farias@h-da.de.}
\thanks{Manuscript received September 28, 2022; revised xxxx, 2022.}}

%
%

\markboth{Journal of \LaTeX\ Class Files,~Vol.xx, No.xx, September~2022}%
{Shell \MakeLowercase{\textit{et al.}}: Bare Demo of IEEEtran.cls for Biometrics Council Journals}
%

\IEEEtitleabstractindextext{%
\begin{abstract}
Non-referential face image quality assessment methods have gained popularity as a pre-filtering step on face recognition systems. In most of them, the quality score is usually designed with face matching in mind. However, a small amount of work has been done on measuring their impact and usefulness on Presentation Attack Detection (PAD). In this paper, we study the effect of quality assessment methods on filtering bona fide and attack samples, their impact on PAD systems, and how the performance of such systems is improved when training on a filtered (by quality) dataset. On a Vision Transformer PAD algorithm, a reduction of 20\% of the training dataset by removing lower quality samples allowed us to improve the BPCER by 3\% in a cross-dataset test.
\end{abstract}

\begin{IEEEkeywords}
Biometrics, Presentation Attack Detection, Image quality
\end{IEEEkeywords}
\centering
}
\maketitle

\IEEEdisplaynontitleabstractindextext
\IEEEpeerreviewmaketitle

\IEEEraisesectionheading{\section{Introduction}\label{sec:introduction}}
\IEEEPARstart{B}{iometric} and/or identity verification systems have numerous commercial and industrial applications in fields diverse such as access controls, video surveillance and user validation, among others. Among the techniques commonly used are fingerprint recognition, ID card validation, and iris and face biometry \cite{Iris-liveness, janier-finger, handbook, handbook-PAD}. Due to the improvement in the quality and availability of capture devices (i.e. smartphones), and the increase in remote image processing, these applications are increasingly available to the general public in uncontrolled environments \cite{selfie}. In this scenario, one of the most relevant challenges for the diffusion and viability of biometric systems is the problem of impersonation. In this case, a subject tries to present fraudulent evidence to the biometric recognition system to be authorised and obtain resources within the system. This problem transversely affects companies, governments and individual users.

Facial Presentation Attack Detection (PAD) techniques on face biometry are used to overcome this challenge. The objective is to differentiate between a genuine biometric reading of a living subject's face (\emph{bona fide}) and a fake one created by the attacker, using, for example, a photo, video, mask or a different substitute for the face of an authorised subject. Presentation attacks tend to be the most common on authentication systems, especially in uncontrolled environments, because they do not necessarily require knowledge on the part of the attacker trying to be authorised. Active research is conducted on this topic due to the difficulty involved in designing an algorithm that generalises well to different sensors, environmental conditions and methods for identity theft attacks. 

\begin{figure}[t]
    \centering
    \includegraphics[width=1\columnwidth]{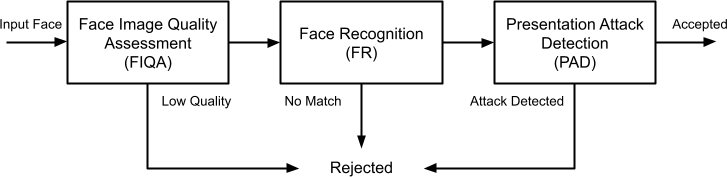}
    \caption{Overview of a traditional remote face authentication system.}
    \label{overview}
\end{figure}

On many real-life remote authentication systems, the user submits an image of his or her face for registering or accessing. In case of accessing, the input face goes to a face recognition algorithm to find a match. This step is done after, or in parallel, with a presentation attack detection system to validate the face. Both steps usually follow a Face Image Quality Assessment (FIQA) that filters out faces not suitable for recognition due to occlusion, bad illumination or heavily rotated faces. Many state-of-the-art face quality algorithms are designed specifically with face recognition as a primary target, tuning their parameters to decrease the matching error. Fig. \ref{overview} shows an overview diagram of a typical remote authentication system.

In this paper, we study the influence of FIQA when applied in conjunction with state-of-the-art face PAD algorithms. Our work proposes the following analysis: first, we analyse the effect of filtering bona fide and attack samples using FIQA algorithms tuned for face recognition. Second, the performance of PAD systems was evaluated when input faces were filtered by their score quality. Finally, we assess the impact on PAD training by discarding low-quality bona fide and attack samples from the training dataset, showing that filtering  by quality on the training dataset can improve the performance of some PAD algorithms even when trained on a lower number of images. 
In summary, the main contributions of this paper are:

\begin{itemize}
    \item We show that a Vision Transformer PAD algorithm \cite{9484333} performs better when presented with higher quality faces even for attack samples, improving its discrimination performance at inference time.
    \item We showed that applying FIQA pre-filtering on an image processing pipeline tends to benefit the selection of bona fide samples over low-quality attacks for most datasets tested.
    \item Training on a reduced dataset, by removing low-quality samples (of both bona fide and attacks) lowers the EER (up to 1\%) and BPCER (up to 4\%) when using ViTranZFAS \cite{9484333} as the PAD algorithm. A similar result is obtained on SSDG \cite{Jia_2020_CVPR_SSDG}.
    
\end{itemize}

\section{RELATED WORK}
Several methods and studies have been done on the effects of applying FIQA algorithms on face recognition systems. A comprehensive study is presented in \cite{10.1145/350790} in which the biometric utility
of face quality data (mostly for visible wavelength images) is discussed on several applications such as face acquisition/enrollment, video frame selection, face detection filtering, conditional image enhancement and model selection, score fusion and presentation attack avoidance. Several FIQA algorithms are summarised and evaluated, highlighting the current predominance of deep learning methods. Finally, future directions and challenges are presented on the topic, especially on the comparability of FIQA algorithms and standardisation. 

A new quality metric, based on vertical edge density, is presented in \cite{7935089} that estimates pose variations on a database of 101 individuals, under different pose angles, illumination and distances, captured on a smartphone device. 

\subsection{Face Presentation Attack Detection}

The role of a presentation attack detection module within a facial recognition system is to prevent unauthorised users from accessing illegally by posing as authorised users. This module checks if the user is real or fake. Generally, if the input image is found to be real, it will proceed to the facial recognition phase; otherwise, it will reject the attempt if it determines that the face is fake. Only recently, commercial facial recognition systems are taking presentation attacks into account as a major issue. It is an important research topic that has gained relevance given the proliferation of facial recognition systems and the technological ease with which physical and digital attacks can be carried out.
In recent years, several articles have been published that comprehensively review state of the art. At a general level, \cite{handbook} is a complete book that explains the problem at the level of general biometrics. More specific reviews, applied to face biometrics can be found at \cite{8229961, 9548404, jimaging6120139, 6990726, lazaro}.
A comprehensive review of algorithms based on deep learning can be found at \cite{yu2021deep}.

\subsubsection{Single Side Domain Generalisation (SSDG)}

SSDG\cite{Jia_2020_CVPR_SSDG} aims to improve the generalisation ability of a presentation attack detection system by learning a compact distribution of real faces' features and a dispersed distribution of the fake ones among domains. A combined feature generation is trained
for both real and fake samples; then the features are fed to a classifier that differentiates between real and fake. Also, the features enter an asymmetric triplet loss computation to make the real samples compact and the fake ones separated by the origin domain. Lastly, the features
of the real samples are used to train a discriminator classifier that aims to modify the feature extractor to learn more generic features across domains, using a Gradient Reversal Layer (GRL).

\subsubsection{ViTranZFAS}

This method \cite{9484333} uses a vision transformer architecture as its backbone, with the last layer replaced by a fully connected layer with one output node and fine-tunes using Binary Cross-Entropy loss (BCE). Our tested method replaces the transformer model with a hybrid model that uses a ResNet-26 Convolutional Neural Network (CNN) to compute the feature map that feeds the Transformer encoder.

\subsubsection{MobileNetv3}

MobileNetv3\cite{mobilenetv3} is a CNN tuned for smartphone CPUs. This CNN adds hard swish activation and squeeze-and-excitation modules, among other changes, to the previous model version, achieving similar accuracy but a considerably faster performance for image classification. In this paper, we used the MobileNetv3 small architecture with pre-trained weights from Imagenet. We modified the net\textquoteright s last layer to be a two-class output instead of the original 1,000 classes.

\subsection{Face Image Quality Assessment}
\label{FIQA_Section}

Face Image Quality Assessment (FIQA) is the process of using face data as input to generate some kind of quality estimate as output. Models can be trained to automatically estimate the quality of a face using man-made scores of the input image (referential FIQA) or without any score as input (non-referential FIQA). In the latter case, the quality metric is usually designed with face comparison in mind; a high-quality face refers to the utility of that face for a face recognition task. In this paper, we evaluate three state-of-the-art non-referential FIQA methods and how they impact the process of presentation attack detection, even when the quality score is not directly related to this task.

\subsubsection{MagFace}

Face picture quality evaluation and face recognition are combined in MagFace\cite{meng2021magface}. The magnitude of the facial recognition feature vector directly relates to quality. Based on its magnitude,
the authors presented a loss function with adaptive margin and regularisation.
The objective of this loss function is to move the challenging samples away from the class centre and the easily recognised examples toward it. Thus, during training, the face utility is inherently learnt by
this loss function. In direct relation to the facial image utility, the magnitude of the feature vector is proportional to the cosine distance to its class centre. A large magnitude denotes a high face utility. Scores range from 0 (lower quality) to 40 (higher quality).

\subsubsection{SER-FIQ}

This method \cite{DBLP:conf/cvpr/TerhorstKDKK20}
relates the robustness of face representations with face quality. Face representations are chosen using sub-networks of randomly changed face models. High-quality face images should be more resilient and have less variability in face representations. A novel idea to quantify face quality based on an arbitrary face recognition model is developed to avoid the usage of unreliable quality labels. The resilience of a sample representation and, consequently, it's quality is measured by examining the embedding variations produced by random sub-networks of a face model. Scores range from 0 (lower quality) to 1 (higher quality).

\subsubsection{SDD-FIQ}

Ou et al. \cite{9577861} present a brand-new unsupervised FIQA technique that uses similarity distribution distance. The technique uses the Wasserstein Distance (WD) between the intra-class and inter-class samples. The FR model uses ResNet-50 trained on the MS1M database to calculate the positive and negative sample distributions. A regression network for quality prediction is trained with Huber loss, using the WD metric as quality pseudo-labels. Scores range from 0 (lower quality) to 100 (higher quality).

\subsection{Face Image Quality applied to PAD}

On FIQA applied to PAD, most works use quality as a means to directly reject fake faces. A software-based fake detection technique is presented in \cite{6671991} that may be applied to various biometric systems (iris, fingerprint and face) to identify various forms of fraudulent access attempts. The approach under consideration aims to increase the security of biometric recognition systems by incorporating anti-spoofing mechanisms in a quick and non-intrusive way using Image Quality Assessment. The method extracts 25 image quality features from an image and can be applied in real-time to distinguish
between real and fake biometric samples.

In order to discern between real and fraudulent face appearances, Fourati et.al.\cite{Fourati} provide a quick and unobtrusive anti-spoofing method based on Image Quality Assessment (IQA) and motion cues. The results produced demonstrated superior performance to cutting-edge
methods. The approach is particularly suited for real-time mobile apps since it considers both dependable robustness and minimal algorithm complexity. 

Chang and Ye \cite{CHANG2022104428} present a face PAD method based on features for multiscale perceptual picture quality evaluation. Specific hand-crafted texture features taken from facial photos are used for spoofing detection. Generalised Gaussian density-based, asymmetric
generalised Gaussian density-based, and top gradient similarity deviation features are the three main models into which the proposed features are classified. A total of 21 multiscale features are gathered for classification using a Support Vector Machine (SVM). Extensive tests
on five benchmark databases as well as a novel dataset, showed that the suggested framework is effective. 

The same authors present in \cite{8354116} a successful method for defending against face spoofing attacks based on perceptual picture quality evaluation features and multi-scale analysis. First, they show that a Blind Image Quality Evaluator (BIQE) is capable of spotting
spoofing attempts. Later, combine the BIQE with an image quality assessment model called Effective Pixel Similarity Deviation (EPSD) to determine the standard deviation of the gradient magnitude similarity map by choosing effective pixels in the image. A multi-scale descriptor
for categorisation is made up of a total of 21 features that were obtained from the BIQE and EPSD. Utilising three current benchmarks, Replay-Attack \cite{replay-attack}, CASIA \cite{casia-surf}, and UVAD \cite{7017526}, extensive research based on intra-class and cross-dataset methods were carried out, showing good performance
compared with state-of-the-art methods.

An extendable multi cues integration framework for face anti-spoofing utilising a hierarchical neural network is proposed to increase the generalisability of face anti-spoofing systems \cite{FENG2016451}.
This framework can combine picture quality cues with motion cues for liveness identification. A liveness feature based on image quality is created using Shearlet. In order to extract motion-based liveness features, dense optical flow is used. Different liveness features can be successfully integrated using a bottleneck feature fusion technique. Three public face anti-spoofing databases were used to evaluate the suggested methodology. 

In \cite{7935080}, real biometric data are distinguished from data used in presentation/sensor spoofing attacks using non-reference image quality criteria. An experimental study demonstrates that true vs. fake iris, fingerprint, and facial data classification are possible
with an average accuracy of 90\% based on a collection of 6 such measures. The target dataset, however, significantly impacts the optimal quality measure (combination) and classification setting, according to this research.

\section{METHOD}

\begin{figure}[t]
    \centering
    \includegraphics[width=1\columnwidth]{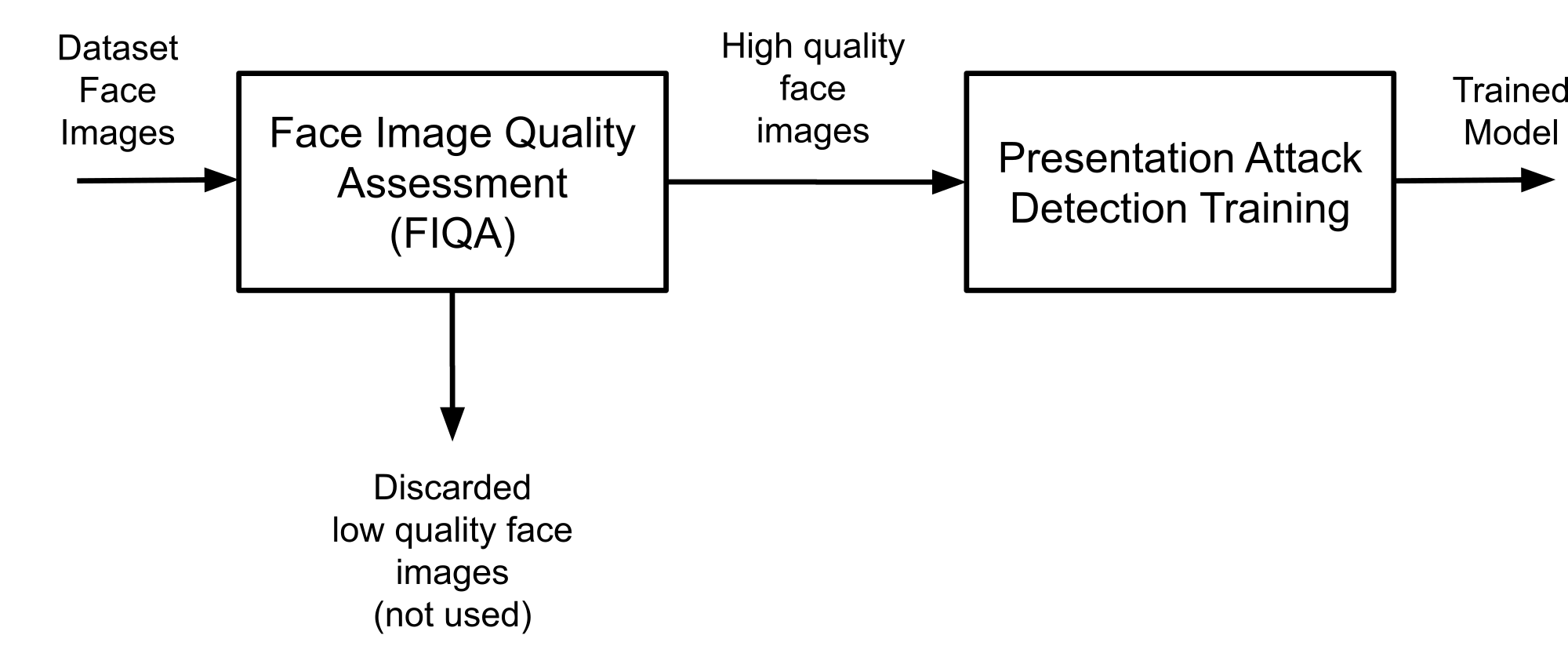}
    \caption{PAD Training with prefiltering of samples by FIQ scores}
    \label{training-pad-fig}
\end{figure}

In this paper, we study the influence of FIQA applied in conjunction with state-of-the-art face PAD algorithms. The following experiments were conducted:
\begin{itemize}
    \item First, we study the distribution of FIQ scores on bona fide and attack samples from several public and proprietary PAD datasets. We evaluated the three FIQA methods described on Section \ref{FIQA_Section}.
    \item Second, the performance of PAD systems was evaluated when input faces were filtered by their score quality at inference time. We evaluated three deep learning-based Face PAD algorithms. Two of them present state-of-the-art results: SSDG\cite{Jia_2020_CVPR_SSDG} and ViTranZFAS\cite{9484333}. The third one is a MobileNetv3\cite{mobilenetv3} based network with a two-class output. We use this method to test the effects on simpler deep-learning networks.
    \item Finally, we assess the impact on PAD training by discarding low-quality bona fide and attack samples from the training dataset for all three PAD methods described above and all three FIQA methods of Section \ref{FIQA_Section}. Figure \ref{training-pad-fig} shows a diagram of the proposed training algorithm.  
\end{itemize}

\subsection{Datasets}
\label{sec_dat_pad}

\begin{figure}[b]
\begin{centering}
\includegraphics[width=1\columnwidth]{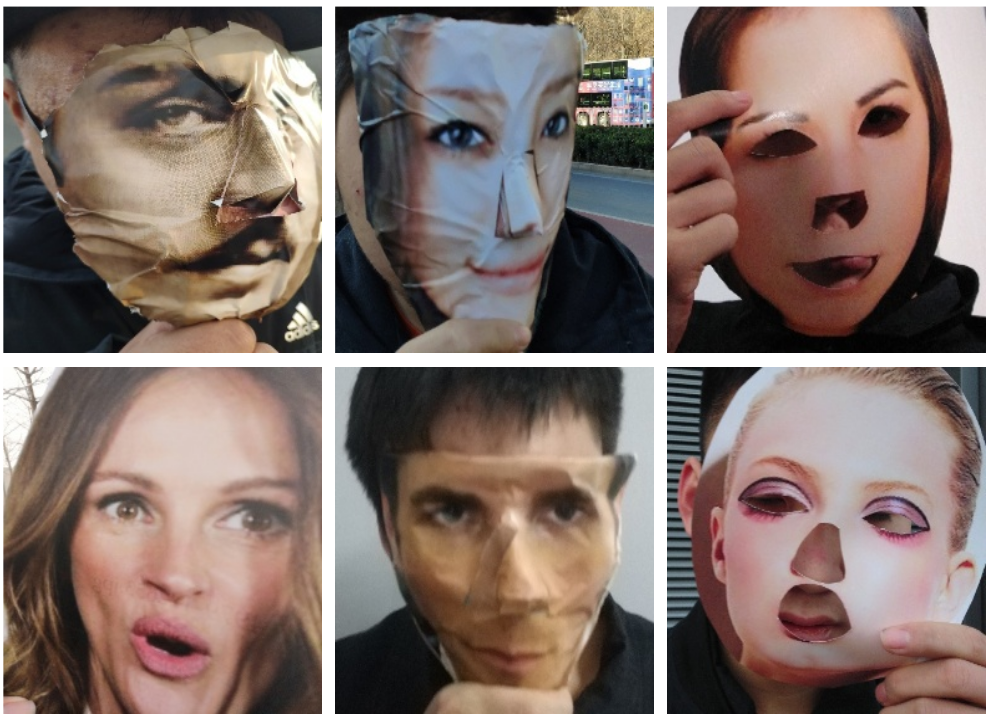}
\par\end{centering}
\caption{\label{fig:celeb_masks}Examples of mask attacks from the CelebA-Spoof dataset.}
\end{figure}

The training was done by merging several face presentation attack datasets to improve generalisation. To enhance the variability of the dataset and increase the number of examples, we included a portion of two face databases that were not related to presentation attack research: the Flickr Face database\cite{8953766} and the UTK Faces database\cite{zhifei2017cvpr}. These datasets have images of regular faces, and print and screen attacks were created artificially using a texture transfer method developed in-house. We also selected a small number of samples \footnote{This file text will be available for reproducibility.} from the CelebA-Spoof dataset\cite{CelebA-Spoof}, for both bona fide and attacks, carefully selecting samples that resemble ``selfie'' images or had mostly frontal faces and attacks. This dataset also has ``mask'' attacks, most of them of low quality. We selected 1,262 masks to be included in out dataset. Fig.\ref{fig:celeb_masks} shows some examples of these images.

\begin{figure}[t]
\begin{centering}
\includegraphics[width=1\columnwidth]{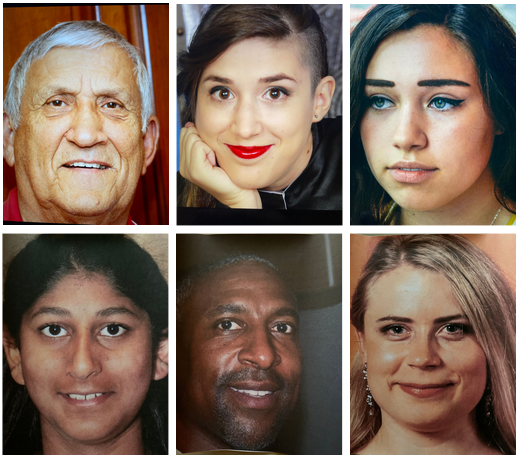}
\par\end{centering}
\caption{\label{fig:Flick-PAD}Flicker-PAD Dataset. Top row: screen attacks.
Bottom row: print attacks.}
\end{figure}

To have high-quality training samples and attacks, we also designed protocols to manually create our own presentation attacks (both printed and screen). Original bona fide samples were taken from the Flickr dataset \cite{8953766} (a different set than the one used for texture transfer) and from our proprietary dataset\footnote{This dataset is sequestered available only for evaluation.} of real-life selfies from users authenticating remotely with their smartphones. We called these datasets Flickr-PAD and Selfies-PAD, respectively. Fig. \ref{fig:Flick-PAD} show some attack samples from the Flickr PAD dataset.

\begin{table*}[t]
\caption{Dataset composition}
\label{train_dataset}

\centering{}%
\begin{tabular}{|c|c|c|c|c|c|}
\hline 
Dataset & Bona fide & Print & Screen & Mask & Comments\tabularnewline
\hline 
\hline 
CASIA-MFSD\cite{6199754} & 279 & 576 & 291 & 0 & 5 frames/video\tabularnewline
\hline 
CelebA-Spoof\cite{CelebA-Spoof} & 2,019 & 1,079 & 1,405 & 1,262 & Manually selected (frontal, selfie like images)\tabularnewline
\hline 
UTK Faces\cite{zhifei2017cvpr} Text.Transfer & 237 & 236 & 236 & 0 & Attacks generated using proprietary texture transfer method \tabularnewline
\hline 
Flickr\cite{8953766} Text.Transfer & 1,580 & 1,580 & 1,579 & 0 &Attacks generated using proprietary texture transfer method\tabularnewline
\hline 
Replay Attack Mobile\cite{replay-mobile} & 2,726 & 3,295 & 1,119 & 0 & 10 frames/video (train+devel)\tabularnewline
\hline 
Flickr\cite{8953766} - PAD & 2,700 & 2,700 & 2,700 & 0 & Proprietary dataset\tabularnewline
\hline 
Selfies - PAD & 2,700 & 5,359 & 4,491 & 0& Proprietary dataset\tabularnewline
\hline 
Total & 12,241 & 14,825 & 11,821 & 1,262 &\tabularnewline
\hline 
\end{tabular}
\end{table*}

Table \ref{train_dataset} shows the composition of this dataset. This dataset has a total of 40,149 images. 35,871 images were used for training purposes. 4,278 images were used as an intra-dataset test for all experiments in this paper, keeping the same class distribution.
Performance of PAD algorithms sharply decreases on out-of-distribution data so we also tested on a cross-dataset testing partition using 2,830 images obtained from the OULU-NPU dataset\cite{oulu}. 

This dataset is distributed as follows: 552 bona fide, 1,155 printed and 1,123 screen attacks. It is essential to highlight that this dataset was not used in training.

\subsection{Training of PAD Algorithms}
All three aforementioned PAD algorithms were implemented on PyTorch
using the Kedro framework \cite{Alam_Kedro_2021}. The same data augmentation scheme was used
on all three algorithms using the albumentation library \cite{info11020125}. Input image
size for all algorithms was $224\times224$ pixels. The training was done on a PC with an Intel i7-9700F CPU, 64 GB of RAM and dual NVIDIA GTX 2080Ti GPUS.

\subsubsection*{ViTranZFAS}

Fine-tuning was performed from pre-loaded weights from the Timm package \cite{rw2019timm}. We used a fixed learning rate of $1e^{-4}$ and weight decay of $1e^{-6}$ with an Adam optimiser. As in the original paper, patch size was kept at $16\times16$ pixels. The training was done with
batch size 64 over 100 epochs.

\subsubsection*{SSDG}

The training was based on pre-trained ImageNet weights, using the following
parameters: weight decay of $5e^{-4}$, the momentum of 0.9, and an initial learning
rate of $1e^{-3}$ (first 150 epochs, then $1e^{-4}$). The $\lambda$
parameters that control the balance between the adversarial loss,
triplet loss and Cross-EntropyLoss were set to 1, 0.5 and 1 respectively.
The training was done on 200 epochs with a batch size of 64.

\subsubsection*{MobileNetv3}

We set an SGD optimiser with a momentum of 0.9, a learning rate of $5e^{-4}$, with Cross-EntropyLoss. The number of workers and batch size were both set to 32. The training was done over 150 epochs. 

\section{EXPERIMENTS AND RESULTS}

\subsection{FIQA effect on filtering presentation attacks}

\begin{figure*}
\includegraphics[width=2\columnwidth]{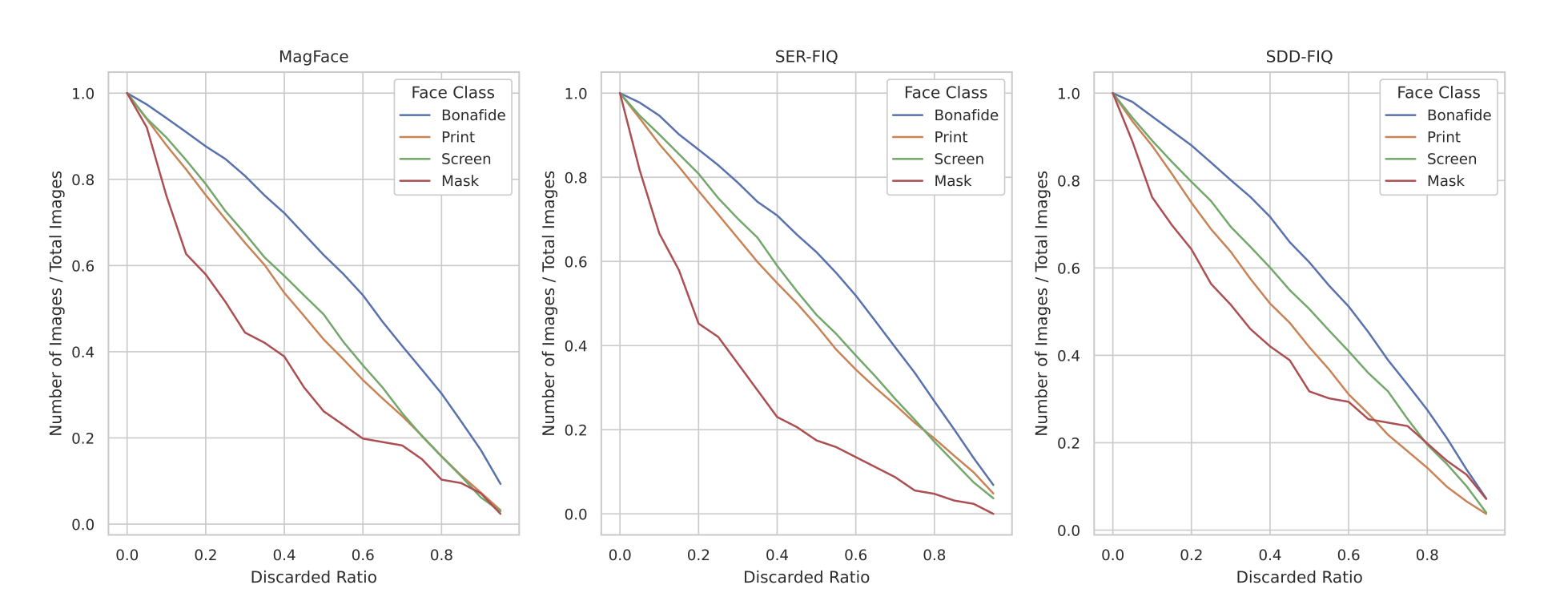}
\caption{Bona fide and presentation attack filtering by quality scores for MagFace, SER-FIQ and SDD-FIQ.}
\label{discarded_images}
\end{figure*}

Quality scores were computed for all 4,278 images of the intra-dataset test for all three FIQ methods used. Then, all images on the lowest X\% with X ranging from 0\% to 95\% (regardless of the class label) were discarded. 
Fig. \ref{discarded_images} shows the ratio of discarded images by class, normalised by the total of images of each class. Results show that bona fide samples are discarded at a lower rate than presentation attacks, with masks being the easiest
to filter out by quality. All three FIQA methods behaved similarly. Even if not designed as a PAD method per se, FIQA pre-filtering on an image processing pipeline tends to benefit the selection of bona fide samples over low-quality attacks.

\begin{figure}[]
\centering
\includegraphics[width=1.0\columnwidth]{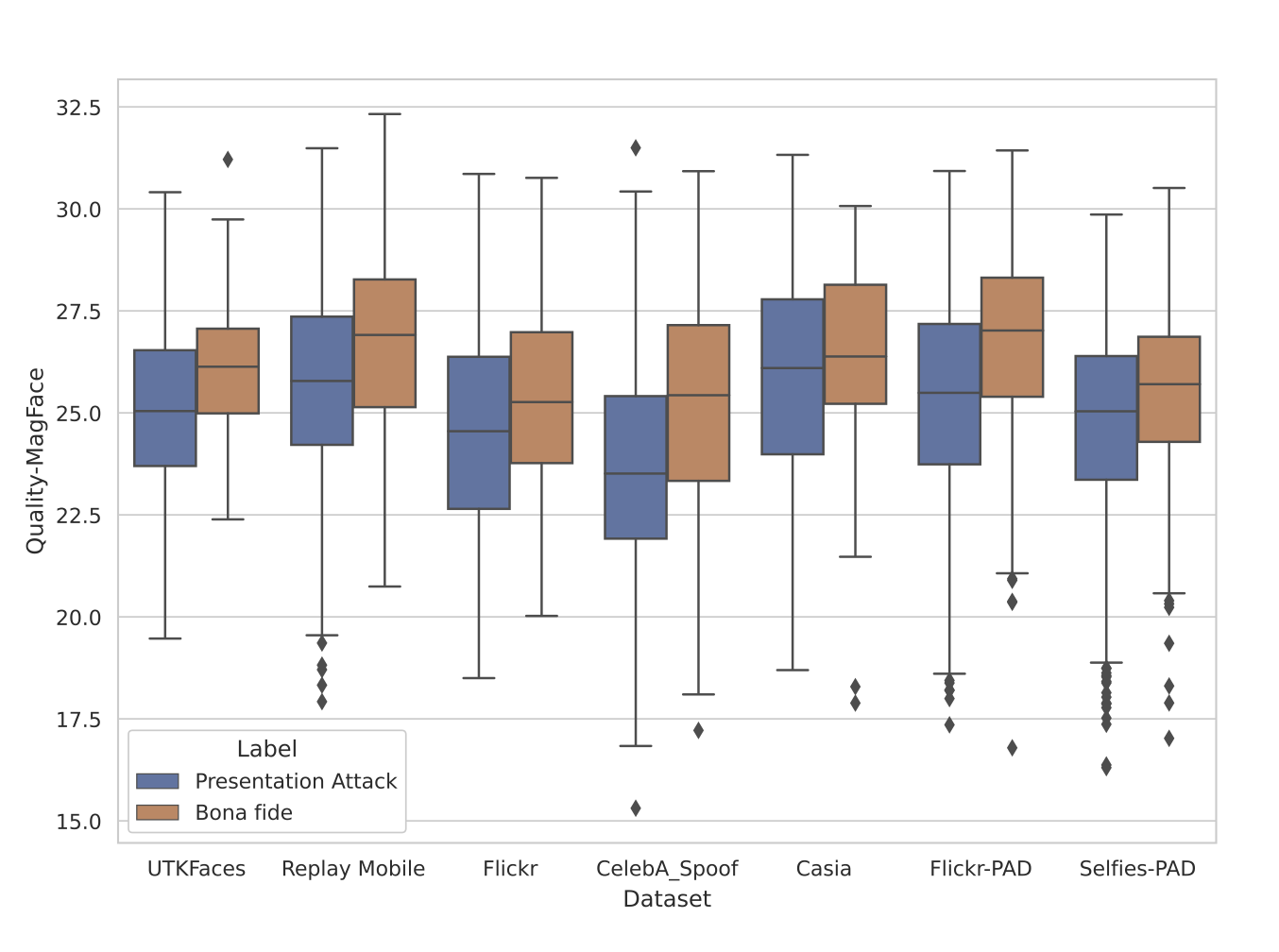}
\caption{Score statistics by original dataset when using MagFace.}
\label{box_magface}
\end{figure}

\begin{figure}[]
\centering
\includegraphics[width=1.0\columnwidth]{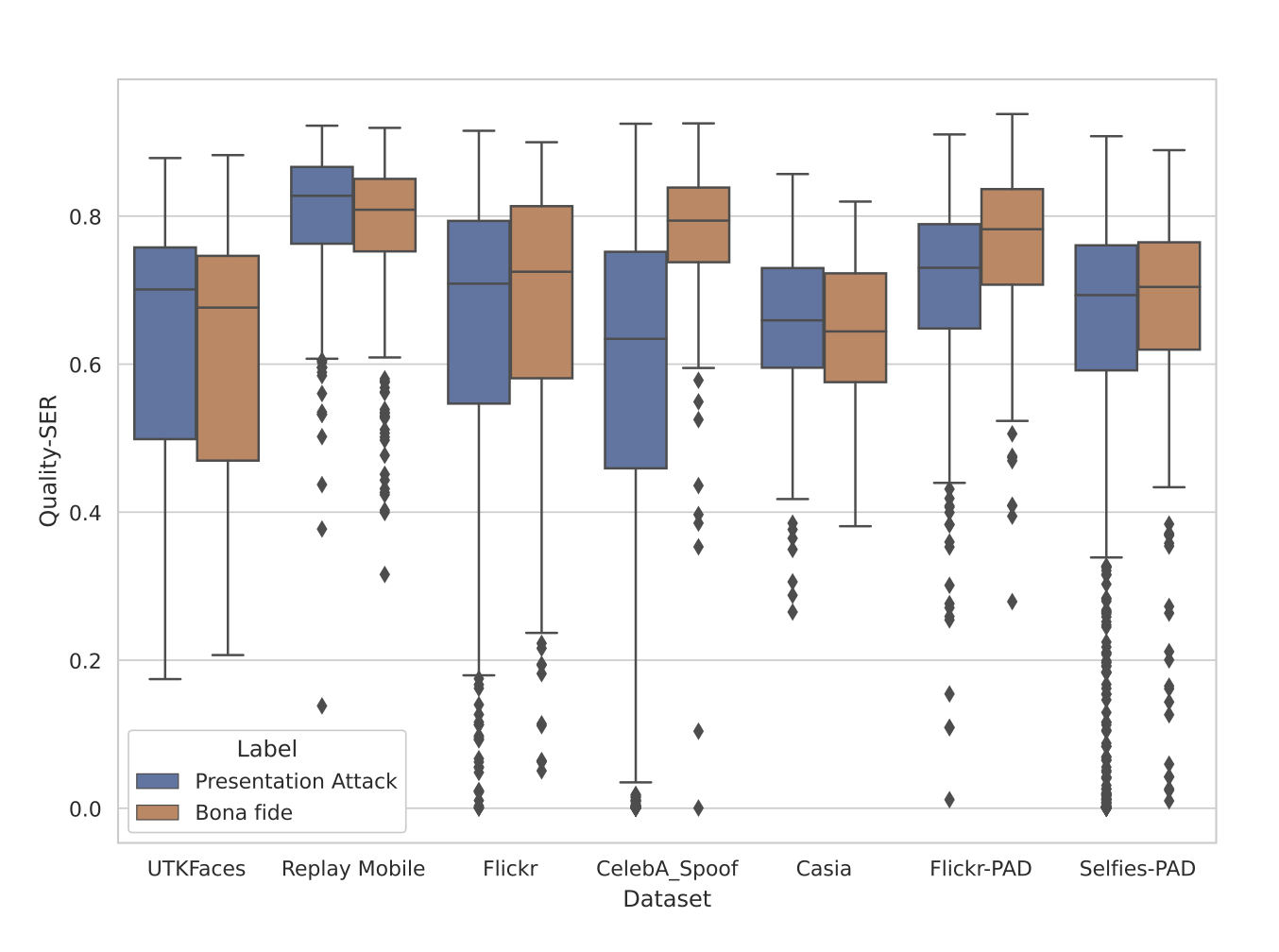}
\caption{Score statistics by original dataset when using SER-FIQ.}
\label{box_ser}
\end{figure}

\begin{figure}[]
\centering
\includegraphics[width=1.0\columnwidth]{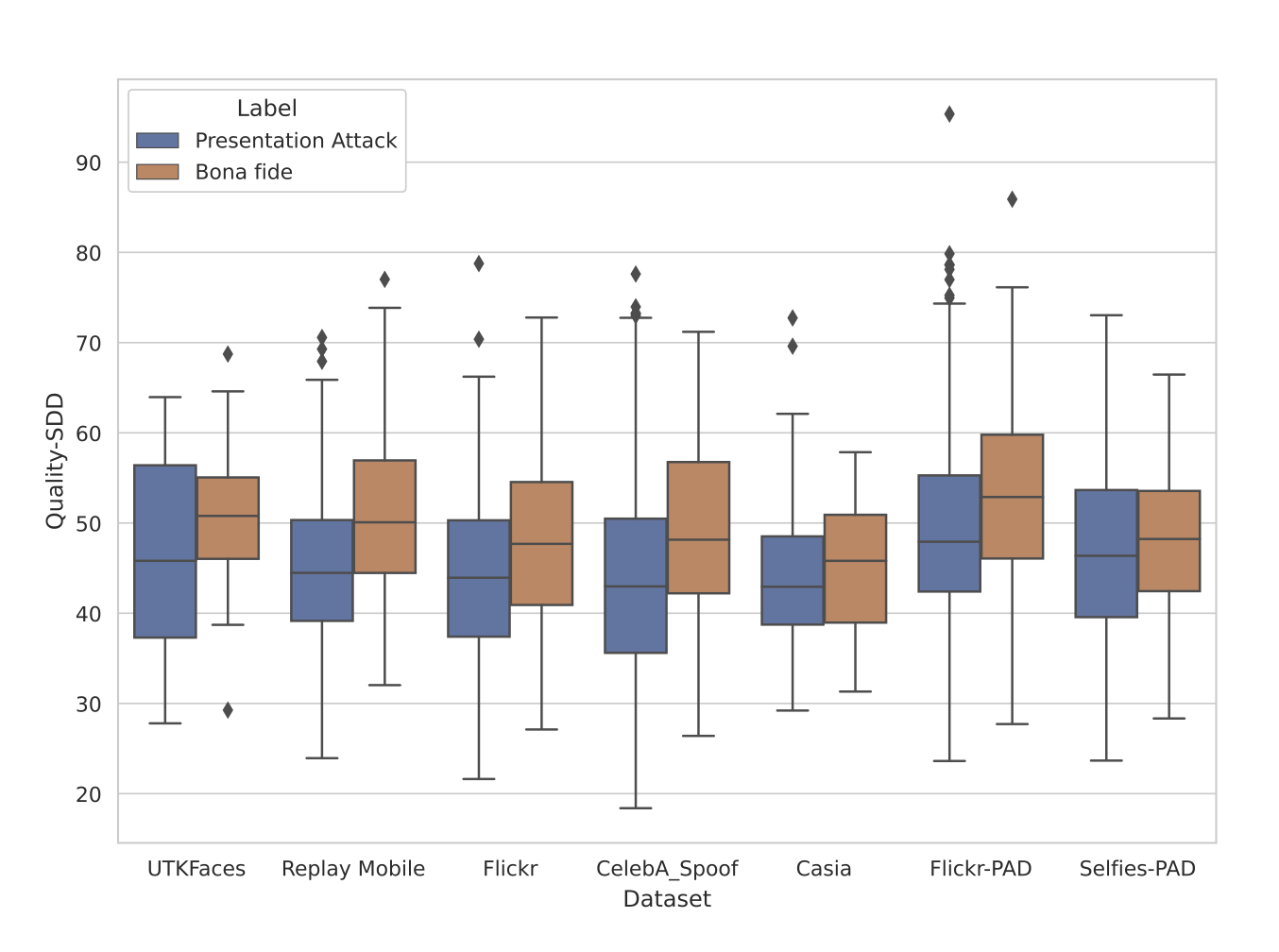}
\caption{Score statistics versus original dataset when using SDD-FIQ.}
\label{box_sdd}
\end{figure}

Figs.\ref{box_magface}, \ref{box_ser} and \ref{box_sdd} show a box plot of the quality scores for each origin on the training dataset for MagFace, SER-FIQ and SDD-FIQ respectively. As expected, attack presentations show a lower score on average than bona fide samples, with some important variation across datasets. 
It should be noted that samples, especially attacks, from the CelebA-Spoof dataset have lower scores than most other datasets. This is expected due to the on-the-wild nature of the images and because many of the presentation attacks are very easily recognised even for the selected fraction of the dataset we are using in this paper. This is unfortunate, given this dataset's massive size and adoption.

\vspace{-0.3cm}

\subsection{PAD performance versus input face quality}

In this experiment, we evaluated the impact on each PAD algorithm when varying the input test images according to their quality score. Images were discarded in steps of 5\% from 0\% to 55\% keeping the higher quality images for each class (so class distribution remains intact across
different discard ratios). The performance of each PAD algorithm was measured according to ISO 30107-3 \footnote{\url{https://www.iso.org/standard/67381.html}} using the Bona fide Presentation Classification Error Rate (BPCER), and Attack Classification Error Rate (APCER) metric defined as (\ref{eq:bpcer}) and (\ref{eq:apcer}).


\begin{equation}\label{eq:bpcer}
    BPCER=\frac{\sum_{i=1}^{N_{BF}}RES_{i}}{N_{BF}}
\end{equation}

\begin{equation}\label{eq:apcer}
    APCER=\frac{1}{N_{PAIS}}\sum_{i=1}^{N_{PAIS}}(1-RES_{i})
\end{equation}

where $N_{BF}$ is the number of bona fide presentations, $N_{PAIS}$ is the number of presentation attacks for a given presentation species and $RES_{i}$ is 1 if the system's response to the i-th attack is classified as an attack and 0 if classified as bona fide. A PAD performance can be reported as a single value of BPCER for a given APCER. For example, $BPCER_{20}$ 
is the BPCER value obtained when the APCER is fixed at 5\%. $BPCER_{10}$  (APCER at 10\%) and $BPCER_{100}$ (APCER at 1\%) are also commonly used.

\begin{figure*}
\includegraphics[width=2\columnwidth]{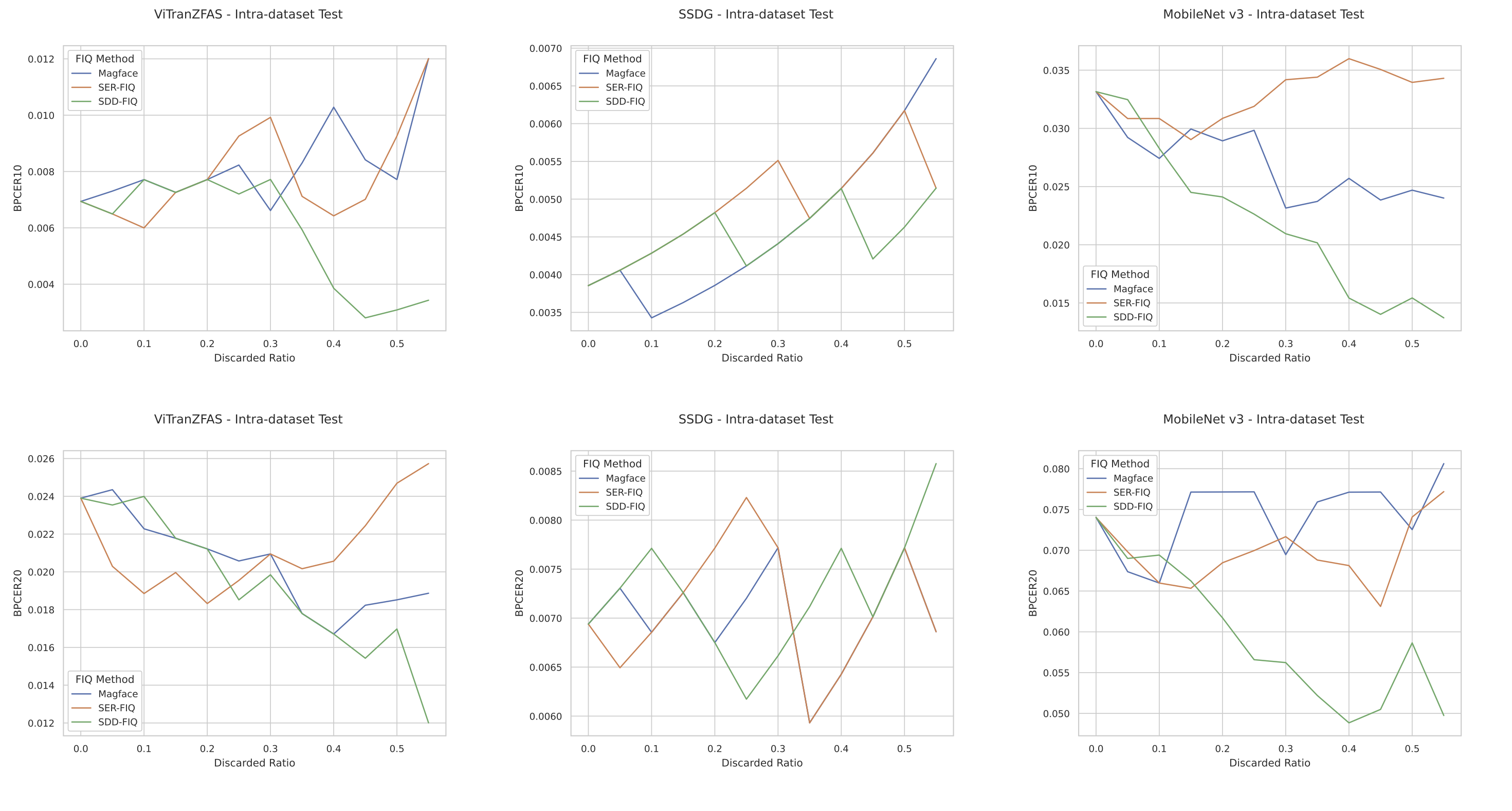}\caption{PAD Performance using BPCER versus Discarded Ratio of images done by Image Quality for the intra-dataset test.} \label{bpcer_intra}
\end{figure*}

\begin{figure*}
\includegraphics[width=2\columnwidth]{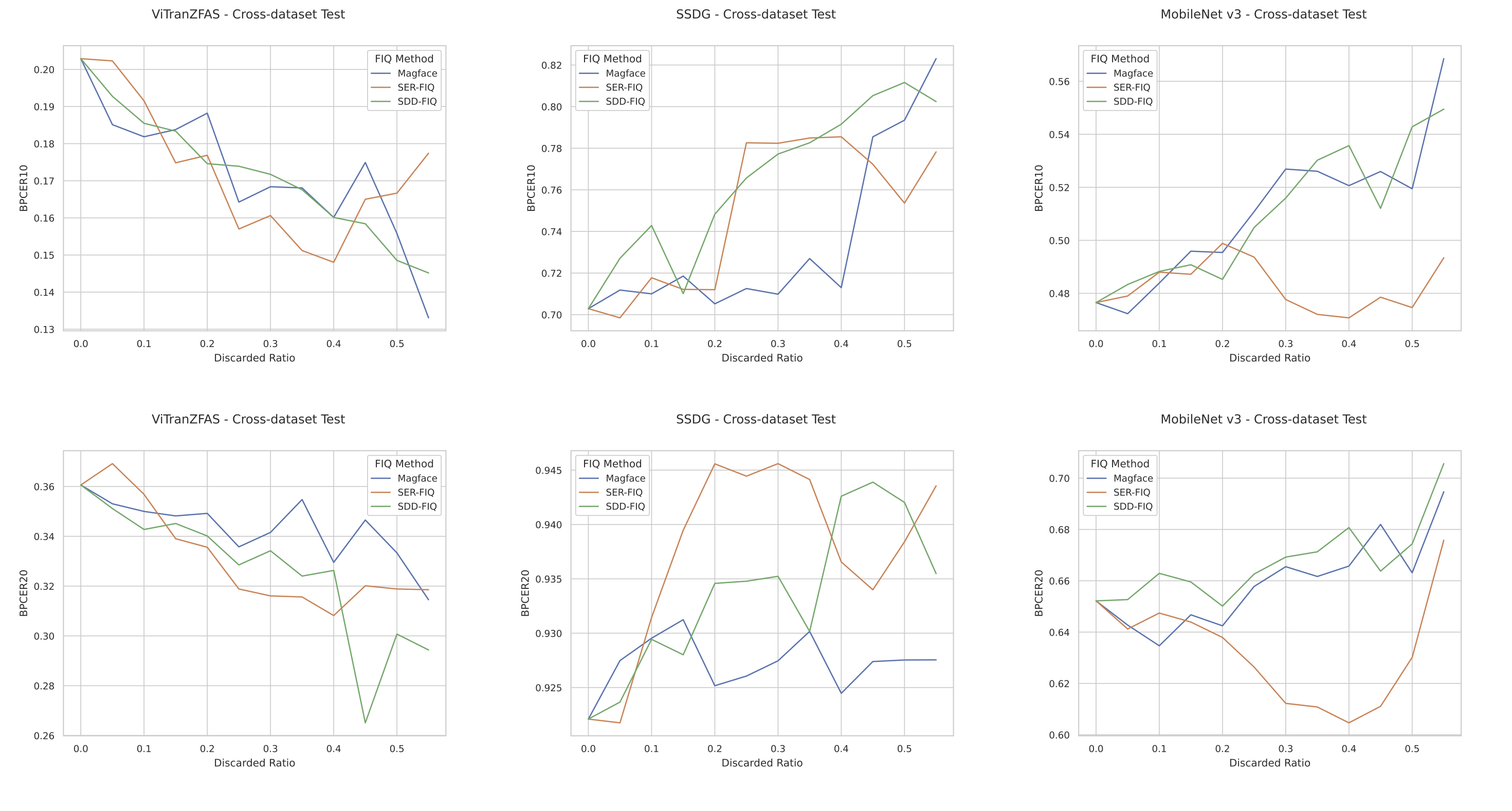}\caption{PAD Performance using BPCER versus Discarded Ratio of images done by Image Quality for the cross-dataset test.} \label{bpcer_cross}
\end{figure*}

Fig. \ref{bpcer_intra} shows the $BPCER_{10}$ and $BPCER_{20}$ metrics on the intra-dataset test for VitranZFAS, SSDG and MobileNetv3. Fig. \ref{bpcer_cross} shows the same experiment evaluated on the cross-dataset test.

\begin{figure}[t]
\begin{centering}
\includegraphics[width=1\columnwidth]{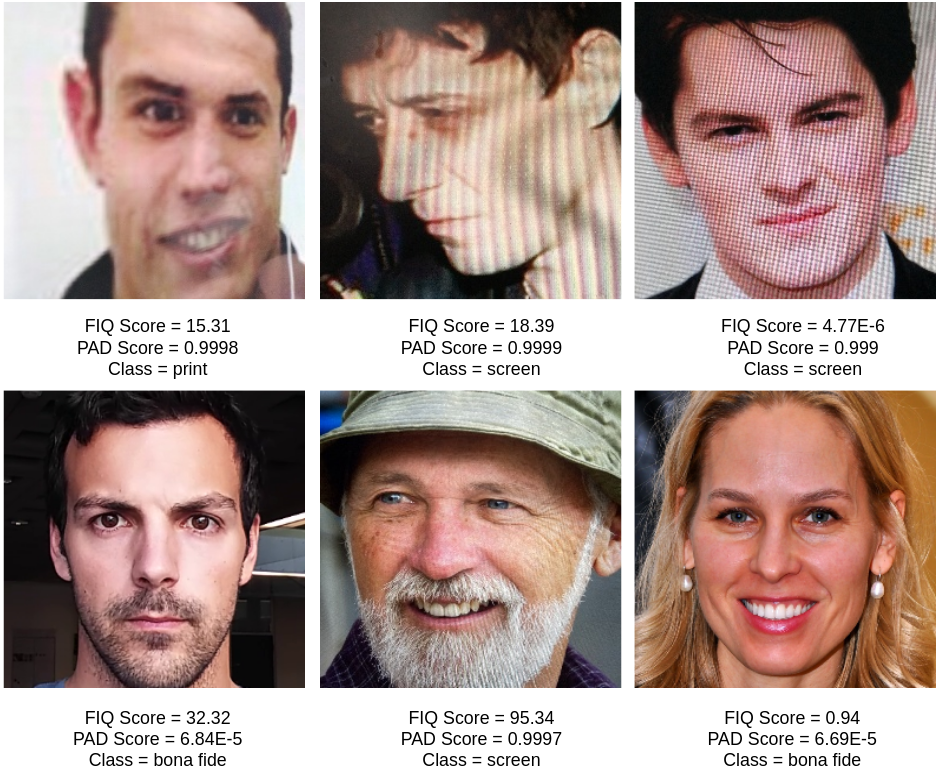}
\par\end{centering}
\caption{\label{fig:Score and Faces} Examples of the lower quality (top) and higher quality images (bottom) for each FIQA method: MagFace (left), SDD (center) and SER (right). The corresponding class and PAD score for the VitranZFAS method is also shown.}
\end{figure}

ViTranZFAS showed a lower error rate on both the intra-dataset (using SDD-FIQ) and cross-dataset (all three FIQA methods) when dealing with higher-quality faces. In the cross-dataset test, the BCPER error decreases by 2\%-4\% when discarding 20\% of the lower-quality images. These results, coupled with the results detailed in the following experiment, seem to indicate than ViTranZFAS performance is improved when presented with higher quality faces even if the overall quality of the attack faces is also improved. 
On the other hand, SSDG and MobileNetv3 showed mixed results, mostly increasing the overall error when the discarded ratio increases. For all three FIQ methods tested, SDD-FIQ showed more consistent performance, followed by MagFace and then SER-FIQ. Note that FIQA these methods are not trained or designed specifically with PAD performance in mind. Figure \ref{fig:Score and Faces} shows the higher and lower quality images for each FIQA method and their corresponding PAD scores for the VitranZFAS algorithm.

\subsection{FIQA filtering of the training dataset}

Face quality scores were computed for the training dataset with all three FIQA methods used in this paper. Then, a fraction of the images (Discarded Ratio) were discarded by sorting them and dropping images with the lowest quality score regardless of sample class (bona fide
or attack). The training was done using the remaining dataset under the same conditions explained in section \ref{sec_dat_pad}. 
Table \ref{pad_training_test} shows the metrics obtained in both the intra-dataset and cross-dataset tests when training with the full dataset and then training with the filtered dataset (EER corresponds to the error rate at the operating point where APCER equals BPCER)

ViTranZFAS showed an improvement of 1.5\% on the EER of the intra-dataset test as well as improved BPCER when training with the MagFace and SDD-FIQ filtered dataset. On the cross-dataset test, improvement is even larger with an improvement close to 4\% on $BPCER_{20}$. Note that training was done with 80\% of the baseline dataset.

When training the SSDG algorithm, a small improvement is shown when training with MagFace on the cross-dataset test. Performance with other FIQ methods is worse than baseline training by a small margin. 
MobileNetv3 training did not show any improvements and cannot compensate for the loss of training images for any of the FIQ methods tested. MagFace showed the best improvements overall with any of the PAD methods. On the other hand, training with the dataset filtered by SER-FIQ did not lower the error metrics for any of the PAD methods tested.

\begin{table*}
\caption{Model Performance versus train dataset Discarded Ratio (DR).}
\label{pad_training_test}
\centering{}%
\begin{tabular}{|l|c|c|c|c|c|c|c|c|}
\hline 
 &  &  & \multicolumn{3}{c|}{Intra-dataset test} & \multicolumn{3}{c|}{Cross-dataset test}\tabularnewline
\hline 
PAD Method & FIQA & DR & $EER${[}\%{]} & $BPCER_{10}${[}\%{]} & $BPCER_{20}${[}\%{]} & $EER${[}\%{]} & $BPCER_{10}${[}\%{]} & $BPCER_{20}${[}\%{]}\tabularnewline
\hline 
\hline 
\multirow{4}{*}{ViTranZFAS} & - & 0 & 3.22 & 0.69 & 2.39 & 13.85 & 20.29 & 36.1\tabularnewline
\cline{2-9} \cline{3-9} \cline{4-9} \cline{5-9} \cline{6-9} \cline{7-9} \cline{8-9} \cline{9-9} 
 & MagFace & 0.2 & 2.00 & 0.31 & 0.62 & 13.04 & \textbf{16.49} & \textbf{33.0}\tabularnewline
\cline{2-9} \cline{3-9} \cline{4-9} \cline{5-9} \cline{6-9} \cline{7-9} \cline{8-9} \cline{9-9} 
 & SER-FIQ & 0.2 & 3.17  & 1.00 & 1.93 & 15.76 & 34.78 & 58.15\tabularnewline
\cline{2-9} \cline{3-9} \cline{4-9} \cline{5-9} \cline{6-9} \cline{7-9} \cline{8-9} \cline{9-9} 
 & SDD-FIQ & 0.2 &  \textbf{1.85} & \textbf{0.08} & \textbf{0.54} & \textbf{12.90} & 19.02 & 35.51 \tabularnewline
\hline 
\multirow{4}{*}{SSDG} & - & 0 & 1.93 & 0.39 & 0.69 & 20.10 & 70.29 & 92.21\tabularnewline
\cline{2-9} \cline{3-9} \cline{4-9} \cline{5-9} \cline{6-9} \cline{7-9} \cline{8-9} \cline{9-9} 
 & MagFace & 0.2 & 2.25 & 0.38 & 1.08 & 18.61 & 64.05 & 87.68\tabularnewline
\cline{2-9} \cline{3-9} \cline{4-9} \cline{5-9} \cline{6-9} \cline{7-9} \cline{8-9} \cline{9-9} 
 & SER-FIQ & 0.2 & 2.78 & 0.46 & 1.93 & 21.47 & 88.59 & 97.46\tabularnewline
\cline{2-9} \cline{3-9} \cline{4-9} \cline{5-9} \cline{6-9} \cline{7-9} \cline{8-9} \cline{9-9} 
 & SDD-FIQ & 0.2 & 2.17 & 0.31 & 0.93 & 20.26 & 71.01 & 92.75\tabularnewline
\hline 
\multirow{4}{*}{MobileNetv3} & - & 0 & 6.19 & 3.31 & 7.40 & 23.20 & 47.65 & 65.22\tabularnewline
\cline{2-9} \cline{3-9} \cline{4-9} \cline{5-9} \cline{6-9} \cline{7-9} \cline{8-9} \cline{9-9} 
 & MagFace & 0.2 & 7.41 & 5.55 & 16.27 & 25.71 & 49.09 & 65.58\tabularnewline
\cline{2-9} \cline{3-9} \cline{4-9} \cline{5-9} \cline{6-9} \cline{7-9} \cline{8-9} \cline{9-9} 
 & SER-FIQ & 0.2 & 7.32  & 5.09 & 19.66 & 22.17 & 49.46 & 67.75\tabularnewline
\cline{2-9} \cline{3-9} \cline{4-9} \cline{5-9} \cline{6-9} \cline{7-9} \cline{8-9} \cline{9-9} 
 & SDD-FIQ & 0.2 & 7.32 & 5.24 & 10.17 & 25.45 & 55.62 & 69.75\tabularnewline
\hline 
\end{tabular}
\end{table*}


\section{Conclusions and Future Work}
In this paper, we evaluated the impact of face quality assessment methods on presentation attack detection systems. We showed that FIQA pre-filtering on an image processing pipeline tends to benefit the selection of bona fide samples over low-quality attacks. Discarding 20\% of the lower quality samples eliminates around 10\% of bona fide but more than 20\% of presentation attack images, more if attacks are of lower quality.
 
ViTranZFAS\cite{9484333} behaves better when presented with high-quality faces (even if they are presentation attacks), improving its attack detection performance. Also, lower EER (up to 1\%) and BPCER (up to 4\%) are obtained when training ViTranZFAS with a filtered training dataset, dropping 20\% of the lower quality samples. This also happens on SSDG \cite{Jia_2020_CVPR_SSDG} when pre-filtering the dataset using MagFace.

For future works, a face quality assessment method, that takes into consideration both presentation attack detection and faces recognition to compute its score, could be implemented to have a more robust face recognition pipeline, especially in real remote applications where fake attempts are more common and easier to make.

Further study is needed on the design of pipelines that include face recognition, presentation attack detection and face quality. Score merging strategies and pipeline order need to be studied to yield better algorithmic performance and cost-effective solutions. Also, the computational efficiency of FIQA algorithms needs to be studied in real-life applications where cost and latency are of uttermost importance.

Finally, this study showed the importance of having more challenging high-quality datasets with carefully made presentation attack images, so PAD algorithms can better learn from harder, more realistic spoofing attempts. There is a lack of publicly available datasets that have the variability, scale and level of crafting for attacks that are required to train PAD systems that can be applied to real applications in the wild with acceptable performance.

\section*{Disclaimer}
This text reflects only the author’s views, and the Commission is not liable for any use that may be made of the information contained therein.

%

\ifCLASSOPTIONcompsoc
  \section*{Acknowledgments}
\else
  \section*{Acknowledgment}
\fi

This work is supported by the European Union’s Horizon 2020 research and innovation program under grant agreement No 883356 and the German Federal Ministry of Education and Research and the Hessen State Ministry for Higher Education, Research and the Arts within their joint support of the National Research Center for Applied Cybersecurity ATHENE and TOC Biometrics R\&D SR-226 cooperation.

\ifCLASSOPTIONcaptionsoff
  \newpage
\fi

{\small

\bibliographystyle{myieee}
\bibliography{biblio}
}

\begin{IEEEbiography}
[{\includegraphics[width=1in,height=1.25in,clip,keepaspectratio]{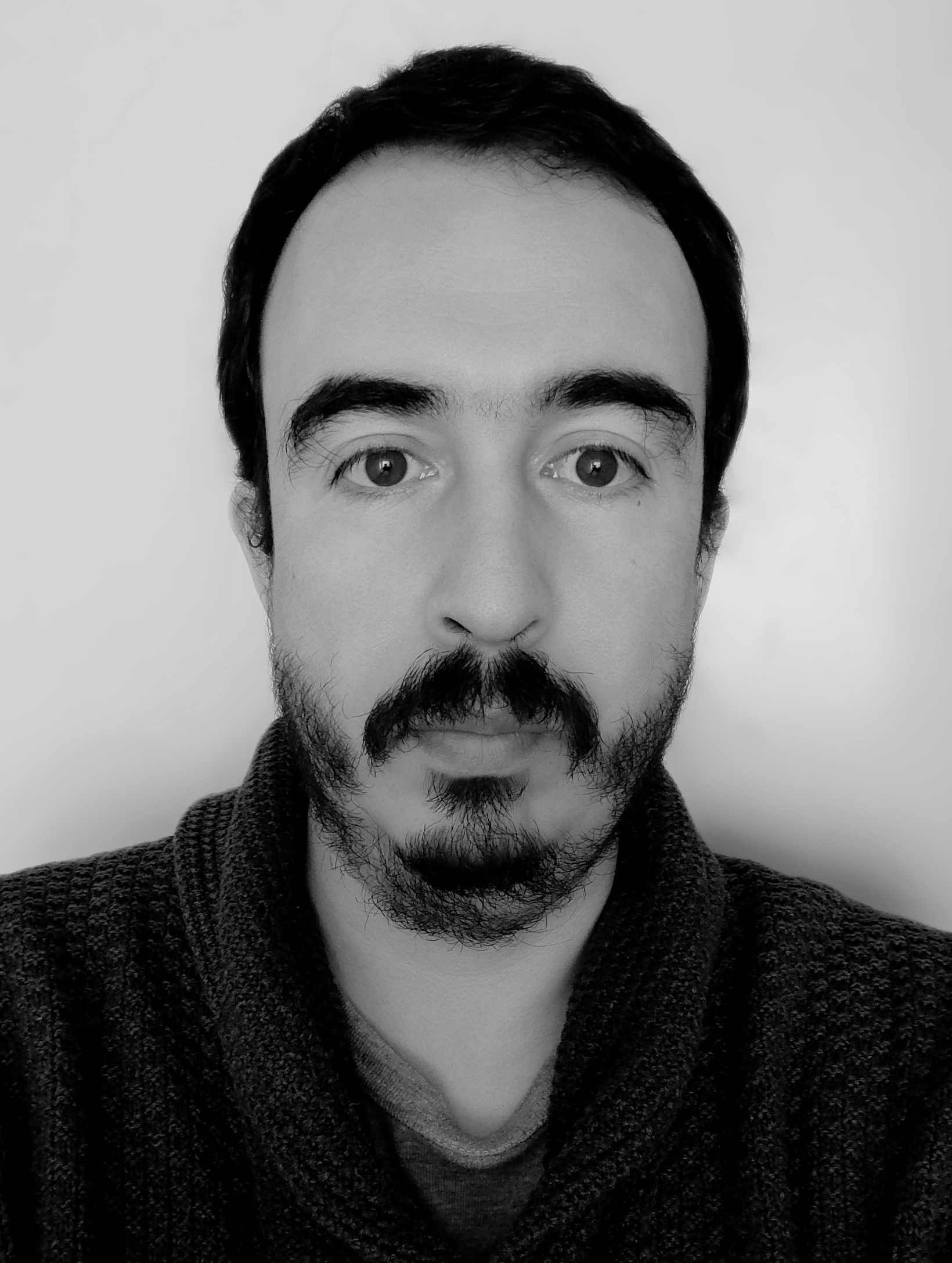}}]{Carlos Aravena} received a P.E. degree in Electrical Engineering from Universidad de Chile in 2009, a M.S. in Electrical Engineering from Universidad de Chile in 2009. He has worked on several applied research projects on face biometrics, object detection and pattern recognition in mining and retail. His main research interests include pattern recognition and deep learning applied to face biometrics, object detection and tracking algorithms. Currently, he is a senior researcher at TOC Biometrics, an R\&D centre.
\end{IEEEbiography}

\begin{IEEEbiography}
[{\includegraphics[width=1in,height=1.25in,clip,keepaspectratio]{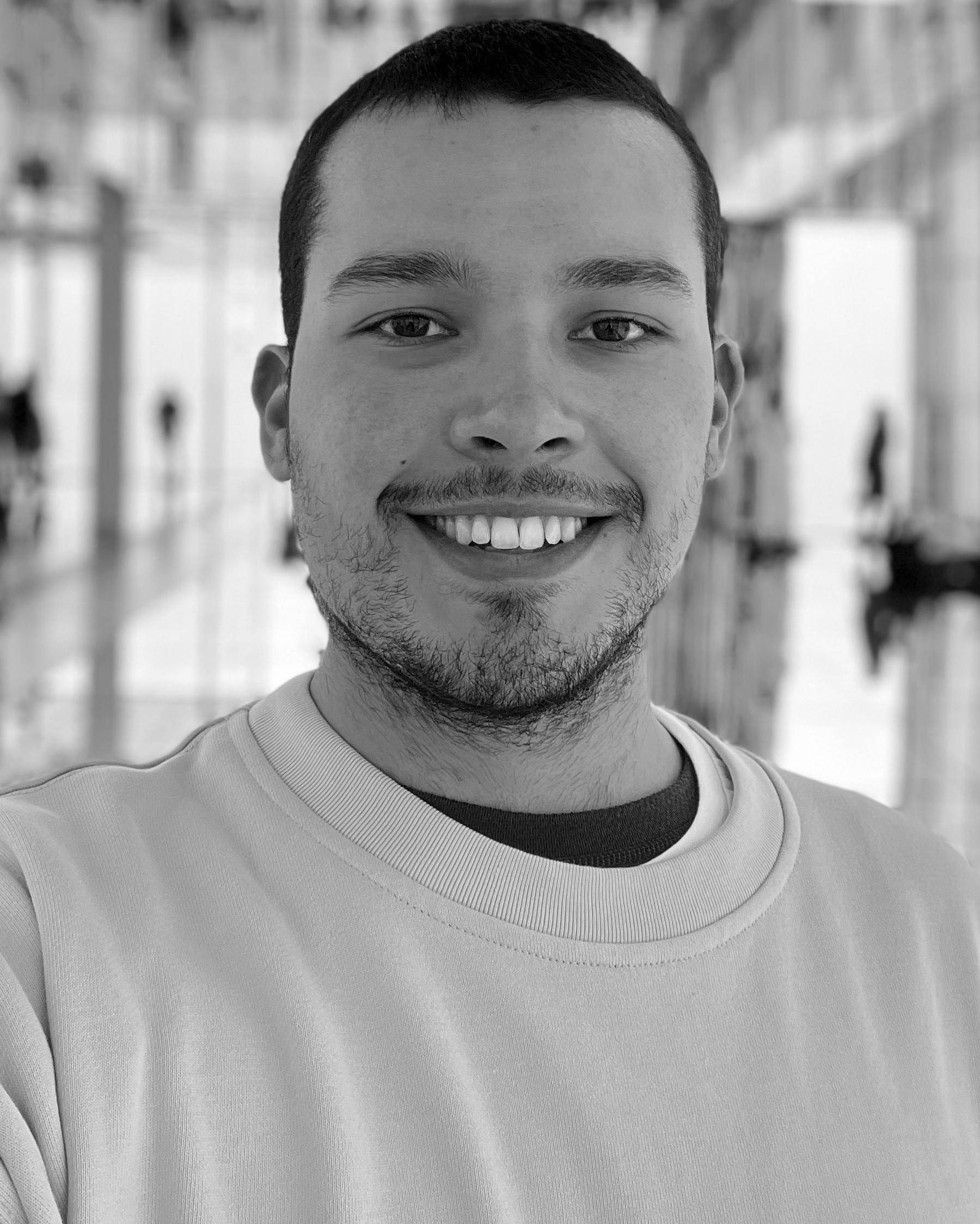}}]{Diego Pasmino} received a P.E. degree in Biomedical Engineering from Universidad de Concepcion in 2020. In addition, His main research interests include pattern recognition, and deep learning applied face biometrics and Presentation Attack detection. Currently, he is a junior researcher at TOC Biometrics, an R\&D centre.
\end{IEEEbiography}

\begin{IEEEbiography}
[{\includegraphics[width=1in,height=1.25in,clip,keepaspectratio]{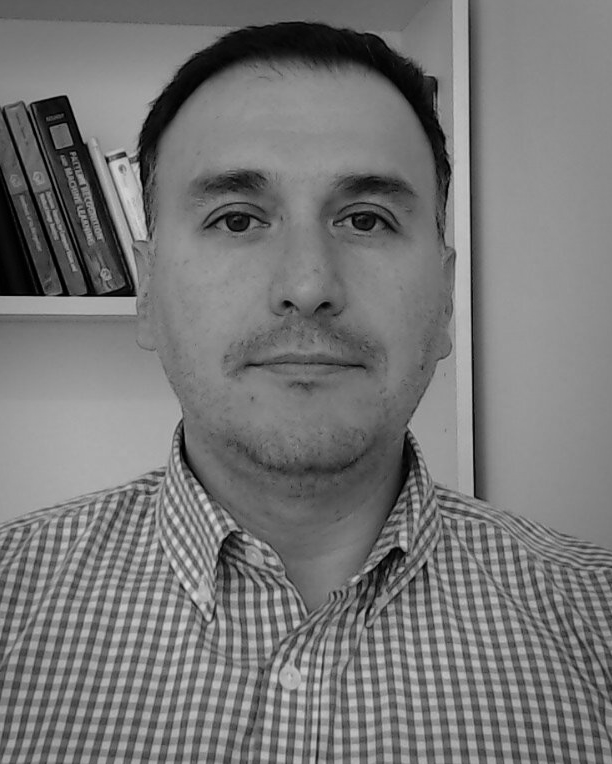}}]{Juan Tapia} received a P.E. degree in Electronics Engineering from Universidad Mayor in 2004, an M.S. in Electrical Engineering from Universidad de Chile in 2012, and a PhD from the Department of Electrical Engineering, Universidad de Chile in 2016. In addition, he spent one year of internship at the University of Notre Dame~(USA). In 2016, he received the award for best PhD thesis. From 2016 to 2017, he was an Assistant Professor at Universidad Andres Bello. From 2018 to 2020, he was the R\&D Director for the area of Electricity and Electricity at Universidad Tecnologica de Chile - INACAP. He is currently a Senior Researcher at Hochschule Darmstadt~(HDA), and R\&D Director of TOC Biometrics. His main research interests include pattern recognition and deep learning applied to iris/face biometrics, vulnerability analysis, morphing, feature fusion, and feature selection.
\end{IEEEbiography}

\begin{IEEEbiography}[{\includegraphics[width=1in,height=1.25in,clip,keepaspectratio]{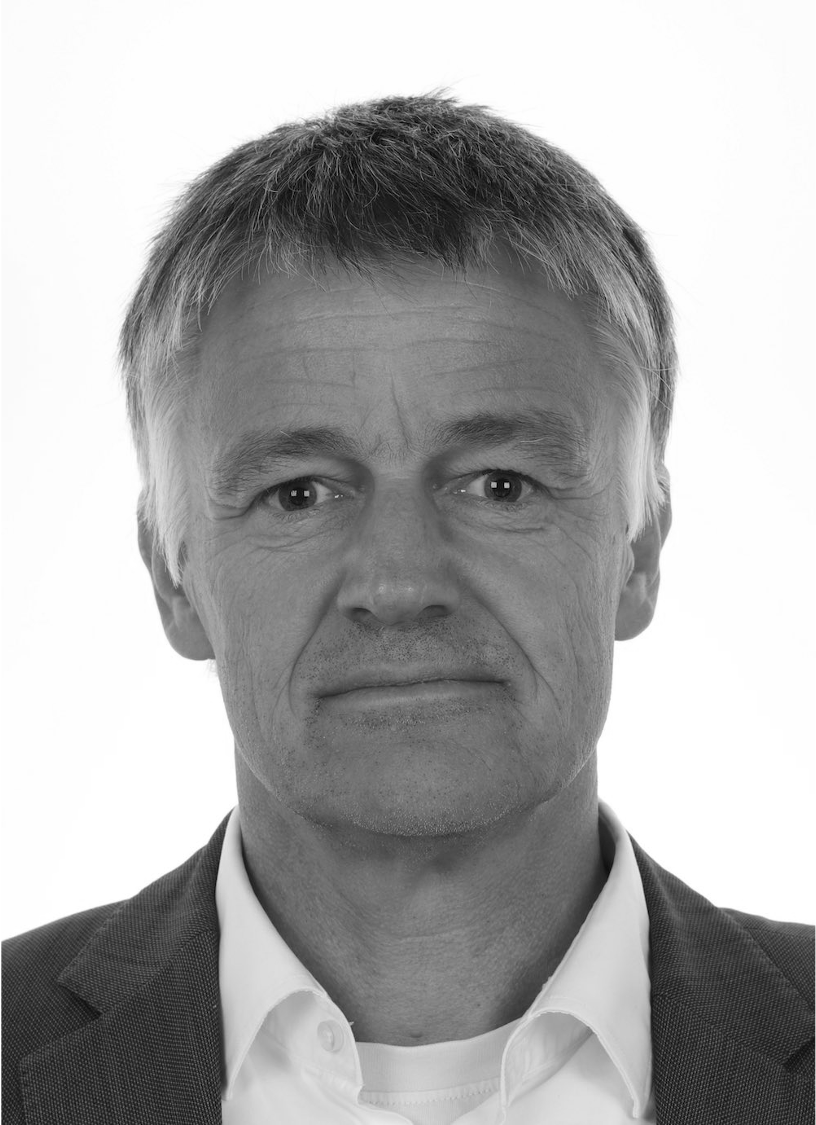}}]{Christoph Busch} is member of the Department of Information Security and Communication Technology (IIK) at the Norwegian University of Science and Technology (NTNU), Norway. He holds a joint appointment with the computer science faculty at Hochschule Darmstadt (HDA), Germany. Further, he lectures the course Biometric Systems at Denmark’s DTU since 2007. On behalf of the German BSI he has been the coordinator for the project series BioIS, BioFace, BioFinger, BioKeyS Pilot-DB, KBEinweg and NFIQ2.0. In the European research program, he was the initiator of the Integrated Project 3D-Face, FIDELITY and iMARS. Further, he was/is a partner in the projects TURBINE, BEST Network, ORIGINS, INGRESS, PIDaaS, SOTAMD, RESPECT and TReSPAsS. He is also a principal investigator in the German National Research Center for Applied Cybersecurity (ATHENE). Moreover, Christoph Busch is a co-founder and member of the board of the European Association for Biometrics (www.eab.org) which was established in 2011 and assembled in the meantime more than 200 institutional members. Christoph co-authored more than 500 technical papers and has been a speaker at international conferences. He is a member of the editorial board of the IET journal.
\end{IEEEbiography}

\end{document}